\documentclass[journal]{IEEEtran}

\usepackage{epsfig}
\usepackage{graphicx}
\usepackage{amsmath}
\usepackage{amssymb}
\usepackage{booktabs}
\usepackage[pagebackref=false,breaklinks=true,letterpaper=true,colorlinks,bookmarks=false]{hyperref}
\usepackage{algorithm}  
\usepackage{algpseudocode}  
\usepackage{multirow}
\usepackage{colortbl}
\definecolor{mygray}{gray}{.9}

\ifCLASSINFOpdf
\else
\fi
\begin{document}
%
\title{Towards Realistic Face Photo-Sketch Synthesis via Composition-Aided GANs}
%
%
%

\author{Jun~Yu,~\IEEEmembership{Senior Member,~IEEE,}
		Xingxin~Xu, 
        Fei~Gao$^{\dag}$, ~\IEEEmembership{Member,~IEEE,}
        Shengjie~Shi, 
        Meng~Wang,~\IEEEmembership{Senior Member,~IEEE,}
        Dacheng~Tao,~\IEEEmembership{Fellow,~IEEE,}
        and~Qingming~Huang,~\IEEEmembership{Fellow,~IEEE}
\thanks{This work was supported in part by the National Natural Science Foundation of China under Grants 61601158, 61971172, 61971339, 61836002, 61702145, 61602136, and 61702143, in part by the China Post-Doctoral Science Foundation under Grant 2019M653563, and in part by the Education of Zhejiang Province under Grants Y201840785 and Y201942162.}
\thanks{Jun Yu, Xingxin Xu, and Shengjie Shi are with the Key Laboratory of Complex Systems Modeling and Simulation, the School of Computer Science and Technology, Hangzhou Dianzi University, 
Hangzhou 310018, China (email: yujun@hdu.edu.cn, 361857031@qq.com, jacobshi777@hotmail.com).}
\thanks{Fei Gao is with the Key Laboratory of Complex Systems Modeling and Simulation, School of Computer Science and Technology, Hangzhou Dianzi University, Hangzhou 310018, China; and the State Key Laboratory of Integrated Services Networks, the School of Electronic Engineering, Xidian University, Xi'an 710071, China (email: gaofei@hdu.edu.cn).}
\thanks{Meng Wang is with the School of Computer Science and Information Engineering, Hefei University of Technology, Hefei 230009, China (email: eric.mengwang@gmail.com)}
\thanks{Dacheng Tao is with The University of Sydney, 6 Cleveland St, Darlington, NSW 2008, Australia (email: dacheng.tao@sydney.edu.au).}
\thanks{Qingming Huang is with the School of Computer Science and Technology, University of Chinese Academy of Sciences, Beijing 100190, China (email: qmhuang@ucas.ac.cn).}
\thanks{$^\dag$ Corresponding author: Fei~Gao, gaofei@hdu.edu.cn.}}

%
%

\markboth{IEEE Transactions on Cybernetics,~Vol.~x, No.~x, xx~2020}%
{Gao \MakeLowercase{\textit{et al.}}: Towards Realistic Face Photo-Sketch Synthesis via Composition-Aided GANs}
%



\maketitle

\begin{abstract}
Face photo-sketch synthesis aims at generating a facial sketch/photo conditioned on a given photo/sketch. It is of wide applications including digital entertainment and law enforcement. Precisely depicting face photos/sketches remains challenging due to the restrictions on structural realism and textural consistency. While existing methods achieve compelling results, they mostly yield blurred effects and great deformation over various facial components, leading to the unrealistic feeling of synthesized images. 
To tackle this challenge, in this work, we propose to use the facial composition information to help the synthesis of face sketch/photo. 
Specially, we propose a novel composition-aided generative adversarial network (CA-GAN) for face photo-sketch synthesis. In CA-GAN, we utilize paired inputs including a face photo/sketch and the corresponding pixel-wise face labels for generating a sketch/photo. Next, to focus training on hard-generated components and delicate facial structures, we propose a compositional reconstruction loss. In addition, we employ a perceptual loss function to encourage the synthesized image and real image to be perceptually similar. Finally, we use stacked CA-GANs (SCA-GAN) to further rectify defects and add compelling details. 
Experimental results show that our method is capable of generating both visually comfortable and identity-preserving face sketches/photos over a wide range of challenging data. In addition, our method significantly decrease the best previous Fr\'{e}chet Inception distance (FID) from 36.2 to 26.2 for sketch synthesis, and from 60.9 to 30.5 for photo synthesis. 
Besides, we demonstrate that the proposed method is of considerable generalization ability. We have made our code and results publicly available: \url{https://fei-hdu.github.io/ca-gan/}.
\end{abstract}

\begin{IEEEkeywords}
Face photo-sketch synthesis, image-to-image translation, generative adversarial network, deep learning, face parsing.
\end{IEEEkeywords}

%
\IEEEpeerreviewmaketitle

\section{Introduction}
%
%
%
%
%

\IEEEPARstart{F}{ace} photo-sketch synthesis refers synthesizing a face sketch (or photo) given one input face photo (or sketch). It has a wide range of applications such as digital entertainment and law enforcement. 
For example, face sketch-synthesis is essential for drawing robots, which draw portraits for human. Besides, face photo-sketch synthesis can significantly boost the accuracy and efficiency of identity verificaiton, in cases where only sketches of criminal suspects are in access \cite{Wang2014A}. 
Ideally, the synthesized photo or sketch portrait should be identity-preserved and appearance-realistic, so that it will yield both high identification accuracy and excellent perceptual quality. 

So far, tremendous efforts have been made to develop facial sketch synthesis methods, both shallow-learning based and deep-learning based \cite{Wang2017RSLCR,Zhang2018Markov,Zhang2019NPGM}.
Especially, inspired by the great success of Generative Adversarial Networks (GANs) \cite{Goodfellow2014GAN} in various image-to-image translation tasks \cite{Isola2017Pix2Pix,Xian2017SurveyGAN}, researchers recently extend GANs for face photo-sketch synthesis \cite{wang2017bpgan,Di2017VAEGAN,Zhang2018IJCAI}. While these methods achieve compelling results, precisely depicting face photos/sketches remains challenging due to the restrictions on structural realism and textural consistency. By carefully examining the synthesized images from existing methods, we observe serious deformations and aliasing defects over the mouth and hair regions. Besides, the synthesized photos/sketches are typically unpleasantly blurred. Such artifacts lead to the unrealistic feeling of synthesized sketches. 

To tackle this challenge, we propose to use the facial composition information to help face photo-sketch synthesis. Specially, we propose to use pixel-wise face labelling masks to character the facial composition. This is motivated by the following two observations. First, pixel-wise face labelling masks are capable of representing the strong geometric constrain and complicated structural details of faces. Second, it is easy to access pixel-wise facial labels due to recent development on face parsing techniques \cite{Liu2015Multi}, thus avoiding heavy human annotations and feasible for practical applications.

Additionally, we propose a composition-adaptive reconstruction loss to focus training on hard-generated components and prevents the large components from overwhelming the generator during training \cite{Lin2017Focal}. In typical image generation methods, the reconstruction loss is uniformly calculated across the whole image as (part of) the objective \cite{Isola2017Pix2Pix}. Thus large components that comprise a vast number of pixels dominate the training procedure, obstructing the model to generate delicate facial structures. However, for face photos/sketches, large components are typically unimportant for recognition (e.g. background) or easy to generate (e.g. facial skin). In contrast, small components (e.g. eyes) typically comprise complicated structures, and thus difficult to generate. To eliminate this barrier, we introduce a weighting factor for the distinct pixel loss of each component, which down-weights the loss assigned to large components. 

We refer to the resulted model as Composition-Aided Generative Adversarial Network (CA-GAN). In CA-GAN, we utilize paired inputs including a face photo/sketch and the corresponding pixel-wise face labelling masks for generating the portrait, and use the improved reconstruction loss for training. Moreover, we use a perceptual loss \cite{Johnson2016Perceptual} based on a pre-trained face recognition network, to further boost the realism of synthesed photos/sketches. Finally, we use stacked CA-GANs (SCA-GAN) for refinement, which proves to be capable of rectifying defects and adding compelling details \cite{Zhang2017StackGAN}. As the proposed framework jointly exploits the image appearance space and structural composition space, it is capable of generating natural face photos and sketches. Experimental results show that our methods significantly outperform existing methods in terms of perceptual quality, and obtain better or comparable face recognition accuracies. We also verify the excellent generalization ability of our new model on faces in the wild. 

In summary, we have made the following contributions: 
\begin{itemize}
\item First, we propose to use the facial composition information to help the synthesis of face sketch/photo, and design a novel generator architecture accordingly. To the best of our knowledge, this is the first work to employ facial composition information in the loop of learning a face photo-sketch synthesis model.
\item Second, we propose a novel compositional loss to focus training on hard-generated components and delicate facial structures, which proves to improve the realism of synthesized photos/sketches. 
\item Third, we employ a perceptual loss to enforce high-frequency and identity constraints on the synthesized images, which encourages the synthesized photos/sketches to be perceptually realistic with preserved identity.
\item Incrementally, we use a stack of our models to further rectify defects and add compelling details, and train it in an end-to-end manner;
\item Our model significantly decrease previous state-of-the-art Fr\'{e}chet Inception distance (FID) from 36.2 to 26.2 for sketch synthesis, and from 60.9 to 30.5 for photo synthesis. Besides, we demonstrate that the proposed method is of considerable generalization ability. We have made our code and results publicly available: \url{https://github.com/fei-hdu/ca-gan}.
\end{itemize}

The rest of this paper is organized as follows. Section \ref{sec:related} introduces related works. Section \ref{sec:method} details the proposed method. Experimental results and analysis are presented in section \ref{sec:experiment}. Section \ref{sec:conclusion} concludes this paper.

\section{Related Work}
\label{sec:related}

\subsection{Face Photo-Sketch Synthesis}
\label{sec:fss}

Tremendous efforts have been made to develop facial photo-sketch synthesis methods, which can be broadly classified into two groups: data-driven methods and model-driven methods \cite{Wang2017Data}.
The former refers to methods that try to synthesize a photo/sketch by using a linear combination of similar training photo/sketch patches \cite{AddRef04, Song2017Stylizing, Gao2012Face, Song2014Real, Pan2012Semi, Wang2009Face}. These methods have two main parts: similar photo/sketch patch searching and linear combination weight computation. The similar photo/sketch searching process heavily increases the time consuming for test.  
Model-driven refers to methods that learn a mathematical function offline to map a photo to a sketch or inversely \cite{Peng2016Multiple, Zhang2015Face, Zhang2016Robust, Wang2013Transductive}. Traditionally, researchers pay great efforts to explore hand-crafted features, neighbour searching strategies, and learning techniques.
However, these methods typically yield serious blurred effects and great deformation in synthesized face photos and sketches.

Recently, a number of trials are made to learn deep learning \cite{Yu2017TCYB} based face sketch synthesis models. For example, 
Zhang et al. \cite{Zhang2016Content} propose to use branched fully convolutional network (BFCN) for generating structural and textural representations, respectively, and then use face parsing results to fuse them together. However, the resulted sketches exists heavily blurred and ring effects.
More recently, inspired by the great success achieved by conditional Ganerative Network (cGAN) \cite{Krizhevsky2012ImageNet,Goodfellow2014GAN,Tembine2019GANTC} in various image-to-image translation tasks \cite{Johnson2016Perceptual,li2019CASI}, researchers extend GANs for face photo-sketch synthesis \cite{wang2017bpgan,Di2017VAEGAN,Zhang2018IJCAI}.
To name a few, 
Wang et al. \cite{wang2017bpgan} propose to first generate a sketch using the vanilla cGANs \cite{Isola2017Pix2Pix} and then refine it by using a post-processing approach termed back projection. 
Di et al. combine the Convolutional Variational Autoncoder and cGANs for attribute-aware face sketch synthesis \cite{Di2017VAEGAN}. However, there are also great deformation in various facial parts. Wang et al. follow the ideas of Pix2Pix \cite{Isola2017Pix2Pix} and CycleGAN \cite{Zhu2017CycleGAN}, and use multi-scale discriminators for generating high-quality photos/sketches \cite{wang2017multgan}. 

For now, a number of works have been proposed to further boost the performance. To name a few, Zhang et al. \cite{Zhang2018IJCAI} embed photo priors into cGANs and design a parametric sigmoid activation function for compensating illumination variations. 
Peng et al. \cite{Peng2019DeepPGM} use a Siamese network to extract deep patch representation and combine it with a probabilistic graphical model for robust face sketch synthesis.
Zhang et al. \cite{Zhang2019TIP} propose a dual-transfer method to improve the face recognition performance. Zhu et al. \cite{Zhu2019ColGAN} and Zhang et al. \cite{Zhang2019MAL} propose to map photos and sketches to a common space, so as to add a consistency constraint to the mappings between paired photo-sketches. 
Zhang et al. \cite{Zhang2019TCYB} propose to use three modules for producing high-quality sketches. Specially they use a U-Net to produce a coarse result, a traditional method to produce fine details for important face components, and a CNN to produce the high-frequency band.

Few exiting methods use the composition information to guide the generation of the face sketch \cite{Zhang2016Content, Zhang2017Compositional} in a heuristic manner. In particular, they try to learn a specific generator for each component and then combine them together to form the entire face. Similar ideas have also been proposed for face image hallucination \cite{Song2017Learning, Huang2017Beyond}. In contrast, we propose to employ facial composition information in the loop of learning to boost the performance.

\subsection{Image-to-image Translation}
\label{sec:cgan}

Our work is highly related to image-to-image translation, which has achieved significant progress with the development of generative adversarial networks (GANs) \cite{Goodfellow2014GAN, Liu2016CoupledGAN} and variational auto-encoders (VAEs) \cite{Makhzani2015VAE}. Among them, conditional generative adversarial networks (cGAN) \cite{Isola2017Pix2Pix} attracts growing attentions because there are many interesting works based on it, including conditional face generation \cite{Kerras2017progressive}, text to image synthesis \cite{Zhang2017StackGAN}, and image style transfer \cite{Chen2017StyleBank}. Inspired by these observations, we are interested in generating sketch-realistic portraits by using stacked cGAN. However, we found the vanilla cGAN  \cite{Isola2017Pix2Pix} insufficient for this task, thus propose to boost the performance by both developing the network architecture and modifying the objective. 

Among existing works, stacked networks have achieved great success in various directions, such as unsupervised image generation \cite{Huang2016Stacked}, unsupervised image-to-image generation \cite{Li2018Unsupervised}, and text-to-image generation \cite{Zhang2017StackGAN}. All of them obtained amazing results. Our stacked GAN is similar to these works, but has the following differences:
1) Previous works only use the noise vector, source image, or text vector as input. In contrast, we use a source image as well as its composition information as input;
Correspondingly, our generators in both stage-I and stage-II contain two encoders for extracting composition and appearance representations, respectively. In contrast, generators in previous stacked GANs contain one single encoder;
2) Previous stacked GANs use global L1 loss and adversarial loss for training. We additionally use compositional L1 loss and perception loss, which significantly improve the quality of generated images; and 
3) In previous stacked GANs, Stage-I GAN is first trained and then fixed while training Stage-II GAN. In our stacked GAN, both Stage-I GAN and Stage-II GAN are trained jointly in an end-to-end manner.

\section{Method}
\label{sec:method}

\subsection{Preliminaries}
\label{sec:preliminary}

The proposed method is capable of handling both sketch synthesis and photo synthesis, because these two procedures are symmetric. In this section, we take face sketch synthesis as an example to introduce our method. 

Our problem is defined as follows. Given a face photo $\mathbf{X}$, we would like to generate a sketch portrait $\mathbf{Y}$ that shares the same identity with sketch-realistic appearance. Our key idea is using the face composition information to help the generation of sketch portrait.
The first step is to obtain the structural composition of a face. Face parsing can assign a compositional label for each pixel in a facial image. We thus employ the face parsing result (i.e. pixel-wise labelling masks) $\mathcal{M}$ as prior knowledge for the facial composition.
The remaining problem is to generate the sketch portrait based on the face photo and composition masks: $\{\mathbf{X}, \mathcal{M}\} \mapsto \mathbf{Y}$. Here, we propose a composition-aided GAN (CA-GAN) for this purpose.
We further employ stacked CA-GANs (SCA-GAN) to refine the generated sketch portraits.
Details will be introduced next.

\subsection{Face Decomposition}
\label{sec:faceparsing}

Assume that the given face photo is $\mathbf{X} \in \mathbb{R}^{m \times n \times d}$, where $m$, $n$, and $d$ are the height, width, and number of channels, respectively. We decompose the input photo into $C$ components (e.g. hair, nose, mouth, etc.) by employing the face parsing method proposed by Liu et al. \cite{Liu2015Multi} due to its excellent performance. For notational convenience, we refer to this model as \texttt{P-Net}. By using P-Net, we get the pixel-wise labels related to 8 components, i.e. two eyes, two eyebrows, nose, upper and lower lips, inner mouth, facial skin, hair, and background \cite{Liu2015Multi}. 

Let $\mathcal{M} = \{\mathbf{M}^{(1)}, \cdots, \mathbf{M}^{(C)} \} \in \mathbb{R}^{m \times n \times C}$ denote the pixel-wise face labelling masks. Here, $\mathbf{M}^{(c)}_{i,j} \in [0, 1], \text{s.t.~} \sum_c{\mathbf{M}^{(c)}_{i,j}}=1 $ denotes the probability pixel $\mathbf{X}_{i,j}$ belongs to the $c$-th component, predicted by P-Net, $c=1, \cdots, C$ with $C=8$. We use soft labels (probabilistic outputs) in this paper. In the preliminary implementation, we also tested our model while using hard labels (binary outputs), i.e. each value $\mathbf{M}^{(c)}_{i,j}$ denotes whether $\mathbf{X}_{i,j}$ belongs to the $c$-th component.
Because it is almost impossible to get absolutely precise pix-wise face labels, using hard labels occasionally yields deformation in the border area between adjacent components.

\begin{figure}
\begin{center}
\includegraphics[width=0.8\linewidth]{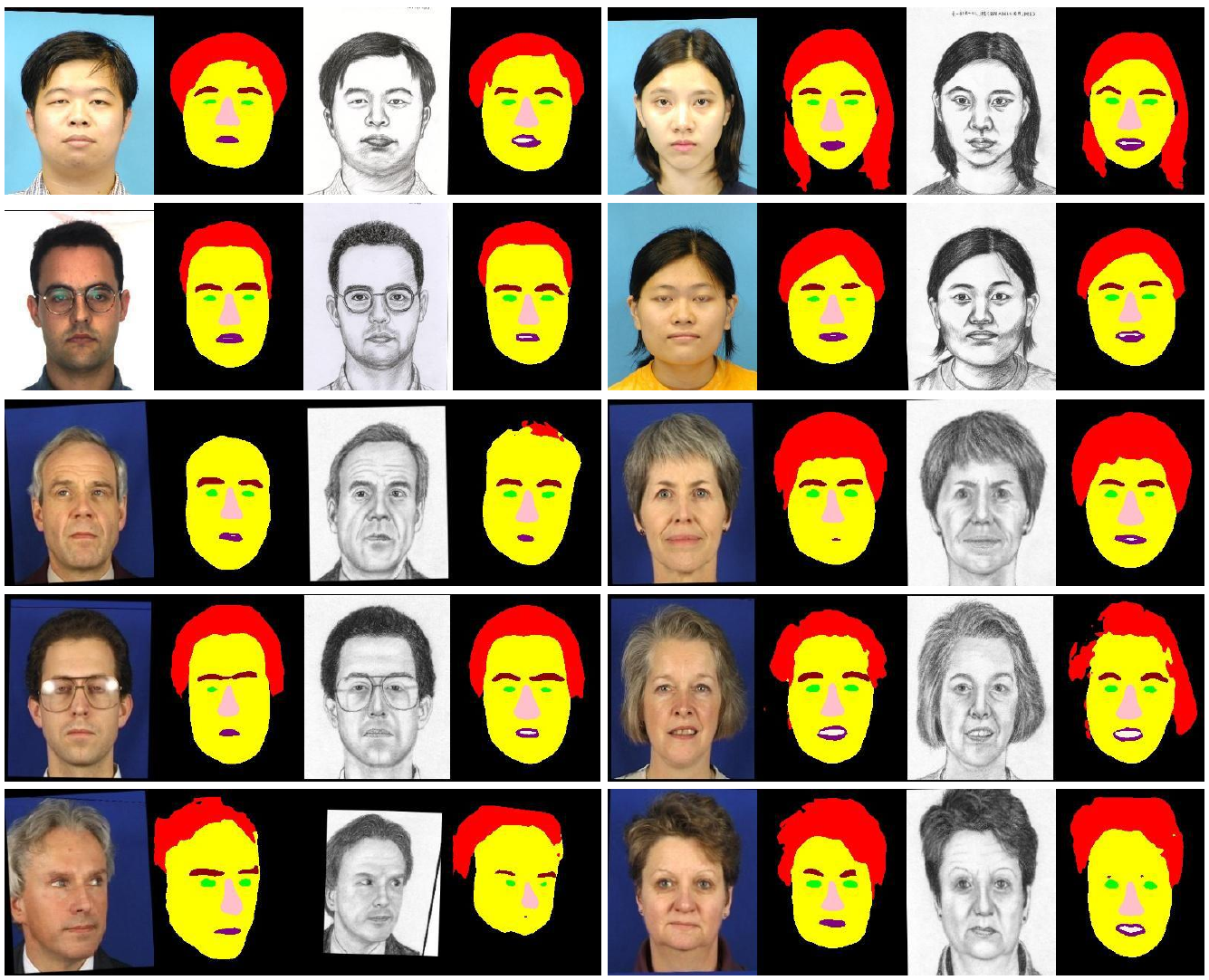}
\end{center}
   \vspace{-0.4cm}
   \caption{Illustration of parsing results produced by P-Net \cite{Liu2015Multi}. For each sample, from left to right are photo, parsing result of photo, sketch, and parsing result of the sketch, sequentially.}
\label{fig:parsing}
   \vspace{-0.4cm}
\end{figure}

\textbf{Notes:} 
We note that an existing face parsing model \cite{Liu2015Multi} is adopted here, as this paper is mainly to explore how to use facial composition information to boost the performance of photo-sketch synthesis. Specially, we here apply the P-Net \cite{Liu2015Multi} pre-trained for face photos to both photos and sketches. 

Although P-Net is not specially designed or learned for sketches, fortunately, we obtain fairly good parsing results. This might due to the fact that P-Net is capable of extracting high-semantic features from a face sketch. Besides, P-Net involves the consistency of two adjacent pixels and Conditional Random Filed (CRF) inference \cite{Liu2015Multi}, both of which characterize local-dependencies inside an image (photo/sketch) and boost the robustness of P-Net.

Some examples are shown in Fig. \ref{fig:parsing}, where facial components are distinguished in colors.  
As shown in Fig. \ref{fig:parsing}, P-Net works fine for most faces, but may fail in cases. Specially, it is hard for P-Net to precisely detect the hair regions for most face photos/sketches. Besides, P-Net may fail to detect eyes (e.g. the example in the forth row, first column) or part of facial skins (e.g. the example in the bottom row, first column). 
Note that face decomposition could more or less introduce errors, however we make no manual intervention or selection to guarantee the claimed overall accuracy. Besides, replacing P-Net with some more advanced face parsing method, e.g. MaskGAN \cite{CelebAMaskHQ}, does improve the synthesis performance. Related code and results have been released on the project page of this work: \url{https://github.com/fei-hdu/ca-gan}.  

In addition, in our CA-GAN and SCA-GAN, the input photo/sketch and its corresponding parsing masks are complementary to each other. Even if the parsing masks are not precise enough, the generator is expected to produce a high-quality face sketch/photo. As will be shown in Section \ref{sec:experiment}, the apparent performance improvement of our methods over cGAN further reflects the effectiveness of the parsing results. 
We expect that an advanced face parsing model will be complementary to our approach, but it is slightly beyond the scope of this paper.

\subsection{Composition-aided GAN (CA-GAN)}
\label{sec:composegan}

In the proposed framework, we first utilize paired inputs including a face photo and the corresponding pixel-wise face labels for generating the portrait.
Second, we propose an compositional reconstruction loss, to focus training on hard-generated components and delicate facial structures.
Next, we employ a perceptual loss function to encourage the synthesized image and real image to be perceptually similar. 

\begin{figure}
\begin{center}
\includegraphics[width=1\linewidth]{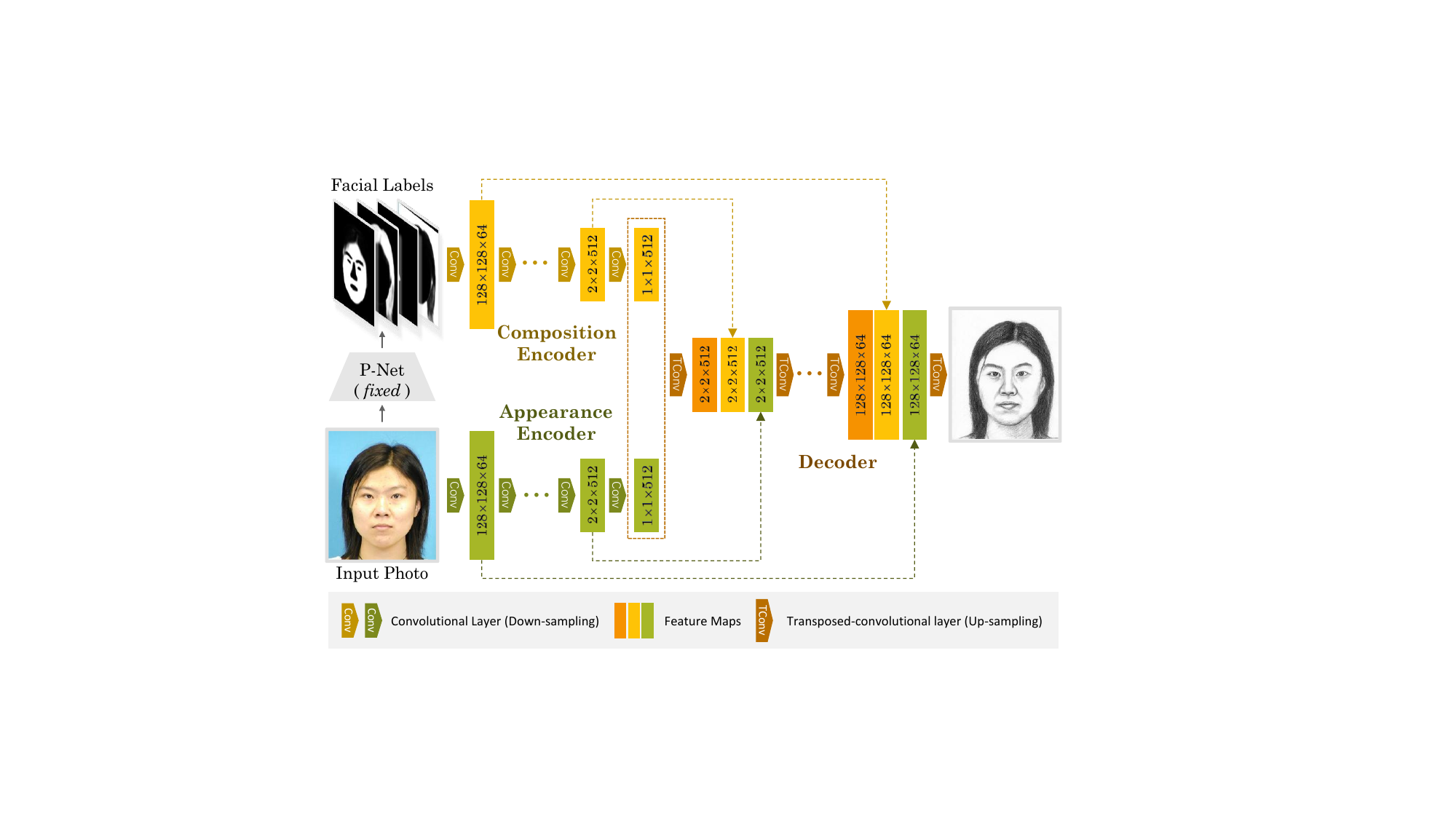}
\end{center}
   \vspace{-0.4cm}
   \caption{Generator architecture of the proposed composition-aided generative adversarial network (CA-GAN).}
\label{fig:cagan}
   \vspace{-0.4cm}
\end{figure}

\subsubsection{Generator Architecture}
\label{sec:garc}

The architecture of the generator in CA-GAN is presented in Fig. \ref{fig:cagan}. For clarity, we illustrate  sizes of feature maps in the format of $width \times height \times the~number~of~channels$. 
In our case, the generator needs to translate two inputs (i.e., the face photo $\mathbf{X}$ and the face labelling masks $\mathcal{M}$) into a single output $\mathbf{Y}$.
Because $\mathbf{X}$ and $\mathcal{M}$ are of different modalities, we propose to use distinct encoders to model them. The corresponding encoders are referred to as \textit{Appearance Encoder} and  \textit{Composition Encoder}, respectively.
The outputs of these two encoders are concatenated at the bottleneck layer for the decoder \cite{Yan2017Skeleton}. In this way, the information of both facial details and composition can be well modelled respectively.
This architecture is different from previous works, where generator typically includes an encoder and a decoder \cite{Isola2017Pix2Pix, Zhu2017CycleGAN, Zhang2017StackGAN}.

The architectures of the encoder, decoder, and discriminator are exactly the same as those used in \cite{Isola2017Pix2Pix} but without dropout, following the shape of a "U-Net". Specifically, we concatenate all channels at layer $i$ in both encoders with those at layer $n-i$ in the decoder. Details of the network will be found in Part \ref{ssec:netarc}. We note that we use the network in \cite{Isola2017Pix2Pix} here because it is a milestone in the image-to-image translation community and has shown appealing results in various tasks. Nevertheless, our proposed techniques are complementary for arbitrary cGANs frameworks. 

In the preliminary experiment, we test the network with one single encoder that takes the cascade of $\mathbf{X}$ and $\mathcal{M}$, i.e. $[\mathbf{X}, \mathbf{M}^{(1)}, \cdots, \mathbf{M}^{(C)}] \in \mathbb{R}^{m \times n \times (d+C)}$, as the input.
This network is the most straightforward solution for simultaneously encoding the face photo and the composition masks. Experimental results show that using this structure decreases the face sketch recognition accuracy by about 2 percent and yield slightly blurred effects in the area of hair.

\subsubsection{Compositional Loss}
\label{sec:comploss}

Previous approaches of cGANs have found it beneficial to mix the GAN objective with a pixel-wise reconstruction loss for various tasks, e.g. image translation \cite{Isola2017Pix2Pix} and super-resolution reconstruction \cite{Johnson2016Perceptual}. Besides, using the normalized $L_1$ distance encourage less blurring than the $L_2$ distance. We therefore use the normalized $L_1$ distance between the generated sketch $\widehat{\mathbf{Y}}$ and the target $\mathbf{Y}$ in the computation of reconstruction loss. We introduce the compositional reconstruction loss starting from the standard reconstruction loss for image generation. 

\paragraph{Global Reconstruction Loss}
In previous works about cGANs, the pixel-wise reconstruction loss is calculated over the whole image. For distinction, we refer to it as \textit{global reconstruction loss} in this paper. Suppose both $\widehat{\mathbf{Y}}$ and $\mathbf{Y}$ have shape $m \times n$. The global reconstruction loss is expressed as:
\begin{equation}
\label{eq:lossglobal}
\mathcal{L}_{L_1, global} (\mathbf{Y}, \widehat{\mathbf{Y}}) = \frac{1}{mn} \Arrowvert \mathbf{Y} - \widehat{\mathbf{Y}} \Arrowvert_1.
\end{equation}

In the global pixel loss, the $L_1$ loss related to the $c^{\text{th}}$ component, $c=1,2,\cdots,C$, can be expressed as:
\begin{equation}
\mathcal{L}_{L_1, global}^{(c)} = \frac{1}{mn} \Arrowvert \mathbf{Y} \odot \mathbf{M}^{(c)} - \widehat{\mathbf{Y}} \odot \mathbf{M}^{(c)} \Arrowvert_1,
    \label{eq:pixlosscmp}
\end{equation}
with $\mathcal{L}_{L_1, global} = \sum_c{\mathcal{L}_{L_1, global}^{(c)}}$. Here, $\odot$ denotes the pixel-wise product operation. 
As all the pixels are treated equally in the global reconstruction loss, large components (e.g. background and facial skin) contribute more to learn the generator than small components (e.g. eyes and mouth).

\paragraph{Compositional Reconstruction Loss}
In this paper, we introduce a weighting factor, $\gamma_c$, to balance the distinct reconstruction loss of each component. Specially, inspired by the idea of balanced cross-entropy loss \cite{Lin2017Focal}, we set $\gamma_c$ by inverse component frequency. Let $\mathbf{1}$ be a $m \times n$ matrix of ones. When we adopt the soft facial labels, $\mathbf{M}^{(c)} \otimes \mathbf{1}$ is the sum of the possibilities every pixel belonging to the $c^\text{th}$ component. Here, $\otimes$ denotes the convolutional operation. If we adopt the hard facial labels, it becomes the number of pixels belonging to the $c^\text{th}$ component. The component frequency is thus $\frac{\mathbf{M}^{(c)} \otimes \mathbf{1}}{mn}$. So we set $\gamma_c = \frac{mn}{\mathbf{M}^{(c)} \otimes \mathbf{1}}$ and multiply it with $\mathcal{L}_{L_1, global}^{(c)}$, resulting in the balanced $L_1$ loss:
\begin{equation}
\mathcal{L}_{L_1, cmp}^{(c)} = \frac{1}{\mathbf{M}^{(c)} \otimes \mathbf{1} } \Arrowvert \mathbf{Y} \odot \mathbf{M}^{(c)} - \widehat{\mathbf{Y}} \odot \mathbf{M}^{(c)}  \Arrowvert_1
	\label{eq:cmploss}
\end{equation}
Obviously, the balanced $L_1$ loss is exactly the normalized $L_1$ loss across the related compositional region.

The \textit{compositional reconstruction loss} is defined as,
\begin{equation}
\mathcal{L}_{L_1, cmp}(\mathbf{Y}, \widehat{\mathbf{Y}}) = \sum_{c=1}^C{\mathcal{L}_{L_1, cmp}^{(c)}}.
\end{equation}
As $\gamma_c$ is broadly in inverse proportion to the component size, it reduces the loss contribution from large components. From the other aspect, it high-weights the losses assigned to small and hard-generated components. Thus the compositional loss focus training on hard components with tiny details, and prevents the vast number of pixels of unimportant component (e.g. background) or easy component (e.g. facial skin) from overwhelming the generator during training.

\paragraph{Compositional Loss} In practice, we use a weighted average of the global reconstruction loss and compositional reconstruction loss:
\begin{equation}
\mathcal{L}_{cmp}(\mathbf{Y}, \widehat{\mathbf{Y}}) = \alpha \mathcal{L}_{L_1, cmp} + (1-\alpha)\mathcal{L}_{L_1, global},
\end{equation}
where $\alpha \in [0, 1]$ is used to balance the global reconstruction loss and the compositional pixel loss. 
We adopt this form in our experiments and set $\alpha=0.7$, as it yields slightly improved perceptual comfortability over the compositional loss. In the following, we refer to the weighted reconstruction loss as \textit{compositional loss}.

\subsubsection{Perceptual Loss}
\label{sec:lid}

In addition, the synthesized image and target image should have similar high-frequency representations and the same identity, which are critical in human perceived quality \cite{Gao2016Biologically, Gao2018Blind}. To this end, we encourage them to have similar feature representations  \cite{Johnson2016Perceptual} as computed by a pre-trained face recognition network, VGGFace \cite{Parkhi15VGGFace}. The perceptual loss is expressed as:
\begin{equation}
\label{eq:lid}
\begin{aligned}
\mathcal{L}_{vggface} = \frac{1}{\lvert \mathcal{S} \rvert} \sum_{l \in \mathcal{S}} \Arrowvert \psi^l(y) - \psi^l(G(x)) \Arrowvert_2,
\end{aligned}
\end{equation}
where $\psi(\cdot)$ denotes the inference process of VGGFace, $\psi^l(\cdot)$ denotes the outputs of the $l$-th layer in the VGGFace; $\mathcal{S}$ is the set of selected layers; $\lvert \mathcal{S} \rvert$ is the number of selected layers. 

Features in different layers of VGGFace contain both high-frequency and identity information. 
Minimizing the perceptual loss for early layers tends to produce images that contain textures indistinguishable from $y$. In contrast, using the loss for higher layers preserves face identity and overall spatial structure \cite{Johnson2016Perceptual}. 
We therefore select both early and high layers, including the \texttt{conv1-1}, \texttt{conv5-1}, and \texttt{conv5-3} layers, in our experiments. 
Using a perceptual loss encourages the output image to be perceptually similar to the target image with the same identity.

\subsubsection{Objective}
\label{sec:objective}

We express the adversarial loss of CA-GAN as \cite{Isola2017Pix2Pix}:
\begin{equation}
\label{eq:ladv}
\begin{aligned}
& \mathcal{L}_{adv}(G, D) = \mathbb{E}_{\mathbf{X}, \mathcal{M},\mathbf{Y} \thicksim p_{data}(\mathbf{X},\mathcal{M},\mathbf{Y})}[\log D(\mathbf{X}, \mathcal{M},\mathbf{Y})] \\
& + \mathbb{E}_{\mathbf{\mathbf{X},\mathcal{M}} \thicksim p_{data}(\mathbf{X},\mathcal{M})}[\log (1-D(\mathbf{X}, \mathcal{M}, G(\mathbf{X}, \mathcal{M})))].
\end{aligned}
\end{equation}
Similar to the settings in \cite{Isola2017Pix2Pix}, we do not add a Gaussian noise $z$ as the input. 

Finally, we use a combination of the adversarial loss, the compositional loss, and the perceptual loss to learn the generator. We aim to solve:
\begin{equation}
\label{eq:objective}
(G^*,D^*) = \arg \min_G \max_D { \mathcal{L}_{adv} + \lambda \mathcal{L}_{cmp} + \gamma \mathcal{L}_{vggface}},
\end{equation}
where $\lambda$ and $\gamma$ are weighting factors. We set them to be 10 and 5, respectively, in the implementation.

\subsection{Stacked Refinement Network}
\label{sec:stackgan}

\begin{figure}
\begin{center}
\includegraphics[width=1\columnwidth]{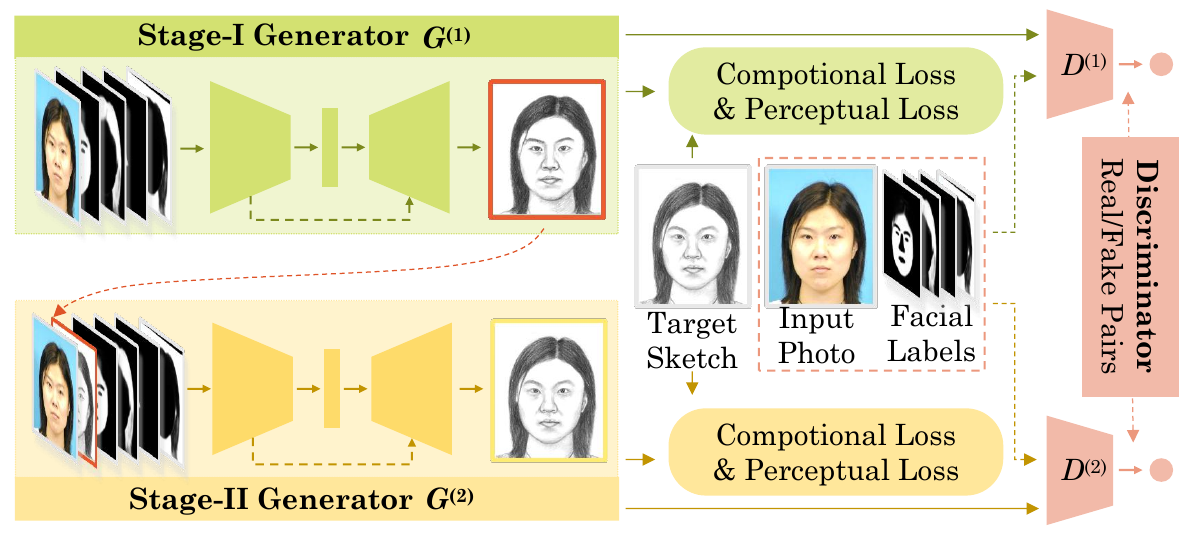}
\end{center}
   \vspace{-0.4cm}
   \caption{Pipeline of the proposed stacked composition-aided generative adversarial network (SCA-GAN).}
\label{fig:stackcagan}
   \vspace{-0.4cm}
\end{figure}

Finally, we use stacked CA-GAN (SCA-GAN) to further boost the quality of the generated sketch portrait \cite{Zhang2017StackGAN}. The architecture of SCA-GAN is illustrated in Fig. \ref{fig:stackcagan}. 

SCA-GAN includes two-stage GANs, each comprises a generator and a discriminator, which are sequentially denoted by $G^{(1)}, D^{(1)}, G^{(2)}, D^{(2)}$. \textit{Stage-I GAN} yields an initial portrait, $\widehat{\mathbf{Y}}^{(1)}$, based on the given face photo $\mathbf{X}$ and pix-wise label masks $\mathcal{M}$. Afterwards, \textit{Stage-II GAN} takes $\{ \mathbf{X}, \widehat{\mathbf{Y}}^{(1)}, \mathcal{M} \}$ as inputs to rectify defects and add compelling details, yielding a refined sketch portrait, $\widehat{\mathbf{Y}}^{(2)}$. Here, $\mathbf{X}$ and $\widehat{\mathbf{Y}}^{(1)}$ are concatenated and input into the appearance encoder of $G^{(2)}$. 

Note that stacking the cGAN also be of benefit. Besides, stacking more than two CA-GANs would further boost the performance. However, using a stack of GANs monotonically increase both the computational complexity and model size, we therefore use a stack of two CA-GANs in the rest of this work, unless otherwise specified. Corresponding analysis will be presented in Part \ref{ssec:exp_stack}. 

\subsection{Network Architectures}
\label{ssec:netarc}

In this work, every photo or sketch is represented in the RGB color space. Following cGAN \cite{Isola2017Pix2Pix}, let $\mathrm{C}i/j$ denote a Convolution-InstanceNorm-LeakyReLU layer with $i$ input channels and $j$ output channels. $\mathrm{TC}i/j$ denotes a TransposedConvolution-InstanceNorm-ReLU. All convolutions are $4\times 4$ spatial filters applied with stride 2. 
Assume that the generator $G^{(k)}$ involves an appearance encoder $E^{(k)}_{a}$, a composition encoder $E^{(k)}_{c}$, and a decoder $Dec^{(k)}$, with $k=1,2$.
SCA-GAN composes of: 
\begin{itemize}
\item \textbf{Encoder}: C$c/64$ $\rightarrow$ C$64/128$ $\rightarrow$ C$128/256$ $\rightarrow$ C$256/512$ $\rightarrow$ C$512/512$ $\rightarrow$ C$512/512$ $\rightarrow$ C$512/512$ $\rightarrow$ C$512/512$. $c$ is sequentially $3$, $8$, $6$, and $8$ for $E^{(1)}_{a}, E^{(1)}_{c}, E^{(2)}_{a}, E^{(2)}_{c}$;

\item \textbf{Decoder}: TC$1024/512$ $\rightarrow$ TC$1536/512$ $\rightarrow$ TC$1536/512$ $\rightarrow$ TC$1536/512$ $\rightarrow$ TC$1536/256$ $\rightarrow$ TC$768/128$ $\rightarrow$ TC$384/64$ $\rightarrow$ TC$192/3$; and

\item \textbf{Discriminator}: C$l/64$ $\rightarrow$ C$64/128$ $\rightarrow$ C$128/256$ $\rightarrow$ C$256/512$ $\rightarrow$ C$512/1$. $l$ is sequentially $14(=3+3+8)$ and $17(=3+3+8+3)$ for $D^{(1)}$ and $D^{(2)}$.
\end{itemize}
The last layer in the decoder is followed by a Tanh function, and the last layer of the discriminator is followed by a Sigmoid function. Besides, InstanceNorm is not applied to the first layer in the encoder. We use leaky ReLUs with slope 0.2 in all the encoders and discriminators. 

\subsection{Optimization}
\label{sec:optim}

To optimize our networks, we alternate between one gradient descent step on $D$, then one step on $G$. We use minibatch SGD and apply the Adam solver. For clarity, we illustrate the optimization procedure of SCA-GAN in Algorithm \ref{alg:optimization}. 
We trained our models on a single Pascal Titan Xp GPU. When we used a training set of 500 samples, it took about 3 hours to train CA-GAN and 6 hours to train SCA-GAN.

\begin{algorithm}    
  \footnotesize
  \caption{Optimization procedure of SCA-GAN (for sketch synthesis).}  
  \label{alg:optimization}
  \begin{algorithmic} 
    \Require  
    	a set of training instances, in form of triplet: \\
     	\{a face photo $\mathbf{X}$, pix-wise label masks $\mathcal{M}$, a target sketch $\mathbf{Y}$ \}; \\
     	iteration time $t=0$, max iteration $T$;
    \Ensure  
        optimal $G^{(1)}, D^{(1)}, G^{(2)}, D^{(2)}$;  
    \State initial $G^{(1)}, D^{(1)}, G^{(2)}, D^{(2)}$; 
    \For{$t=1$ to $T$}  \\
1. Randomly select one training instance: \\
~~~~ \{ a face photo $\mathbf{X}$, pix-wise label masks $\mathcal{M}$, a target sketch $\mathbf{Y}$. \} \\
2. Estimate the initial sketch portrait: \\
~~~~$\widehat{\mathbf{Y}}^{(1)} = G^{(1)}(\mathbf{X}, \mathcal{M}) $ \\
3. Estimate the refined sketch portrait: \\
~~~~$\widehat{\mathbf{Y}}^{(2)} = G^{(2)}(\mathbf{X}, \mathcal{M}, \widehat{\mathbf{Y}}^{(1)}) $ \\
4. Update $D^{(1)}$: \\
~~~~$D^{(1)*} = \arg \min_{D^{(1)}} { \mathcal{L}_{adv}(G^{(1)}, D^{(1)}) }$  \\
5. Update $D^{(2)}$: \\
~~~~$D^{(2)*} = \arg \min_{D^{(2)}} { \mathcal{L}_{adv}(G^{(2)}, D^{(2)}) }$  \\
6. Update $G^{(1)}$: \\
~~~~$G^{(1)*} = \arg \max_{G^{(1)}} { \mathcal{L}_{adv}(G^{(1)}, D^{(1)}) + \lambda \mathcal{L}_{L_1}(\mathbf{Y}, \widehat{\mathbf{Y}}^{(1)})}$  \\
7. Update $G^{(2)}$: \\
~~~~$G^{(2)*} = \arg \max_{G^{(2)}} { \mathcal{L}_{adv}(G^{(2)}, D^{(2)}) + \lambda \mathcal{L}_{L_1}(\mathbf{Y}, \widehat{\mathbf{Y}}^{(2)})}$ 
    \EndFor 
  \end{algorithmic} 
\end{algorithm}

\section{Experiments}
\label{sec:experiment}

In this section, we will first introduce the experimental settings and then present a series of empirical results to verify the effectiveness of the proposed method.  

\subsection{Settings}
\label{sec:settings}

\subsubsection{Datasets}
\label{sec:datasets}

We conducted experiments on two widely used and public available datasets: the CUHK Face Sketch (CUFS) dataset \cite{Ref13} and the CUFSF dataset \cite{Ref31}. The composition of these datasets are briefly introduced below.
\begin{itemize}
\item The CUFS dataset consists of 606 face photos from three datasets: the CUHK student dataset \cite{Ref2} (188 persons), the AR dataset \cite{Ref23} (123 persons), and the XM2VTS dataset \cite{Ref24} (295 persons). For each person, there are one face photo and one face sketch drawn by the artist. 
\item The CUFSF dataset includes 1194 persons \cite{Ref32}. In the CUFSF dataset, there are lighting variation in face photos and shape exaggeration in sketches. Thus the CUFSF dataset is very challenging. For each person, there are one face photo and one face sketch drawn by the artist. 
\end{itemize}

\textit{Dataset partition.} There are great divergences in the experimental settings among existing works. In this paper, we follow the most used settings presented in \cite{Wang2017RSLCR}, and split the dataset in the following ways. For the CUFS dataset, 268 face photo-sketch pairs (including 88 pairs from the CUHK student dataset, 80 pairs from the AR dataset, and 100 pairs from the XM2VTS dataset) are selected for training, and the rest are for testing. For the CUFSF dataset, 250 pairs are selected for training, and the rest 944  pairs are for testing. 

\textit{Preprocessing.} Following existing methods \cite{Wang2017RSLCR}, all these face images (photos and sketches) are geometrically aligned relying on three points: two eye centers and the mouth center. The aligned images are cropped to the size of $250\times{200}$. 
In the CUFS and CUFSF datasets, landmarks of each face photo/sketch are released. For the face images beyond these datasets (Part \ref{sec:crossdataset}), we use MTCNN \cite{Zhang2014MTCNN} for landmark detection. MTCNN is pre-trained and released by the corresponding authors, and has been widely used in face-related tasks. 

In the proposed method, the input image should be of fixed size, e.g. $256 \times 256$. In the default setting of \cite{Isola2017Pix2Pix}, the input image is resized from an arbitrary size to $256 \times 256$. However, we observed that resizing the input face photo will yield serious blurred effects and great deformation in the generated sketch \cite{wang2017bpgan} \cite{wang2017multgan}. In contrast, by padding the input image to the target size, we can obtain considerable performance improvement (as will be seen in Part \ref{ssec:ablation_preproc}). 
We therefore use zero-padding for cGAN, CA-GAN, SCA-GAN, as well as their model variants across all the experiments.

\subsubsection{Criteria}
\label{sec:criteria}
In this work, we choose the \textit{Fr\'{e}chet Inception distance} (FID) to evaluate the realism and variation of synthesized photos and sketches \cite{Heusel2017FID,Lucic2017Are}. 
FID measures the Earth-Mover Distance (EMD) between the distribution of generated samples and that of the ground-truth samples, in the feature space. In this paper, the 2048-dimensional feature of the Inception-v3 network pre-trained on ImageNet is used \cite{szegedy2016rethinking}.
Lower FID values mean closer distances between synthetic and real data distributions. FID has been widely used in image generation tasks and shown highly consistency with human perception. In our experiments, we use all the test samples to compute the FID. 

We additionally adopt the \textit{Feature Similarity Index Metric} (FSIM) \cite{Zhang2011FSIM} between a synthesized image and the corresponding ground-truth image to objectively assess the quality of the synthesized image. 
In FSIM, the phase congruency (PC) and the image gradient magnitude (GM) is employed as features, and the feature similarity between a test image and its corresponding reference image is employed as the quality index. 
Notably, although FSIM works well for evaluating quality of natural images and has become a prevalent metric in the face photo-sketch synthesis community, it is of low consistency with human perception for synthesized face photos and sketches \cite{Wang2016Evaluation}. 

Finally, we statistically evaluate the face recognition accuracy while using the ground-truth photo/sketche as the probe image and synthesized photos/sketches as the images in the gallery. 
\textit{Null-space Linear Discriminant Analysis} (NLDA) \cite{Ref28} is employed to conduct the face recognition experiments. 
NLDA is a face recognition method and is derived from Linear discriminant analysis (LDA) for solving the small sample size problem. In the experiment section, we use “NLDA” to denote the face recognition accuracy by using NLDA.
We repeat each face recognition experiment 20 times by randomly splitting the data and report the average accuracy.

\bigskip We use the proposed architecture for both sketch synthesis and photo synthesis. In the following context, we present a series of experiments:
\begin{itemize}
\item First, we perform ablation study on face photo-sketch synthesis on the CUFS dataset (see Part \ref{sec:ablation}); 
\item Second, we perform face photo-sketch synthesis on the CUFS and CUFSF datasets and compare with existing advanced methods (see Part \ref{sec:photo2sketch} and Part \ref{sec:sketch2photo}); 
\item Third, we conduct experiments on faces in the wild to verify whether the proposed method is robust to lighting and pose variations (see Part \ref{sec:crossdataset}); and 
\item Finally, we verify that the proposed techniques speed up and stabilize the training procedure (see Part \ref{sec:loss}).
\end{itemize}
Our code and results are publicly available at: \url{https://github.com/fei-hdu/ca-gan}.


\subsection{Ablation Study}
\label{sec:ablation}

\begin{table*}
\centering
\caption{Ablation study on the CUFS dataset. Our baseline is cGAN with a face photo/sketch as input. The best index in each column is shown in \textbf{boldface}. $\downarrow$ indicates lower is better, while $\uparrow$ higher is better. In model variants, $\mathbf{X}$ denotes the input face photo, $\mathcal{M}$ the face composition,  $\mathcal{L}_{cmp}$ the compositional loss, $\mathcal{L}_{vggface}$ the perceptual loss $\mathcal{L}_{vggface}$, and \textit{stack} the stacked refinement.}
\footnotesize
\label{tab:ablation}
\begin{tabular}{c|cccccc|ccc|ccc}
\toprule
\multicolumn{7}{c|}{\textit{Model Variants}}					&	\multicolumn{3}{c|}{\textit{Sketch Synthesis}}					&	\multicolumn{3}{c}{\textit{Photo Synthesis}}					\\	
Remarks	& Preprocessing	&	Input $\mathbf{X}$	&	Input $\mathcal{M}$	&	$\mathcal{L}_{cmp}$	&	$\mathcal{L}_{vggface} $	&	Stack	&	FID$\downarrow$	&	FSIM$\uparrow$	&	NLDA$\uparrow$	&	FID$\downarrow$	&	FSIM$\uparrow$	&	NLDA$\uparrow$	\\	
\midrule
	& Resizing &	$\checkmark$	&	-	&	-	&	-	&	-	&	90.8 	&	62.1	&	48.2 	&	132.4 	&	65.6	&	16.9 	\\
cGAN	& Zero-Padding &	$\checkmark$	&	-	&	-	&	-	&	-	&	43.2	&	71.1	&	95.5	&	117.6	&	74.8	&	89.0	\\	
	& Zero-Padding &	-	&	$\checkmark$	&	-	&	-	&	-	&	43.8	&	69.2	&	85.5	&	103.3	&	77.0	&	94.1	\\	
	& Zero-Padding &	$\checkmark$	&	$\checkmark$	&	-	&	-	&	-	&	40.5	&	71.3	&	95.2	&	81.1	&	78.0	&	98.8	\\	
	& Zero-Padding &	$\checkmark$	&	$\checkmark$	&	$\checkmark$	&	-	&	-	&	39.7	&	71.2	&	95.6	&	81.1	&	78.6	&	98.6	\\	
CA-GAN	& Zero-Padding &	$\checkmark$	&	$\checkmark$	&	$\checkmark$	&	$\checkmark$	&	-	&	36.1 
&	71.3 &	95.8 &	41.3	&	78.6	&	98.5 	\\	

\midrule
stack-cGAN	& Zero-Padding &	$\checkmark$	&	-	&	-	&	-	&	$\checkmark$	&	36.6	&	71.2	&	95.3	&	104.3	&	75.5	&	88.0	\\	
SCA-GAN	& Zero-Padding &	$\checkmark$	&	$\checkmark$	&	$\checkmark$	&	$\checkmark$	&	$\checkmark$	&	\textbf{34.2}	&	\textbf{71.6}	&	95.7	&	\textbf{40.3} & \textbf{79.5}	&	\textbf{99.4} 	\\	
\bottomrule	
\end{tabular}
\vspace{-0.2cm}
\end{table*}

\begin{table*}
\centering
\caption{Performance on face sketch-synthesis on the CUFS dataset and CUFSF dataset. The best and second best indices in each line are shown in \textbf{boldface} and \underline{underline} format, respectively. $\downarrow$ indicates lower is better, while $\uparrow$ higher is better. Here, we compare the proposed methods with a number of existing advanced methods, including MRF \cite{Wang2009Face}, MWF \cite{MWF}, SSD \cite{Song2014Real}, MrFSPS \cite{Peng2016Multiple}, RSLCR\cite{Wang2017RSLCR}, BFCN \cite{BFCN}, MRNF \cite{Zhang2018Markov}, DGFL \cite{Zhu2017Deep}, BP-GAN \cite{wang2017bpgan}, and cGAN \cite{Isola2017Pix2Pix}.}
\label{tab:pfm_sketch}
\begin{tabular}{c|c|ccccc|ccc|cccc}
\toprule		
\textit{Criterion} &	\textit{Dataset} & \multicolumn{5}{c|}{\textit{Traditional methods}}	& \multicolumn{3}{c|}{\textit{Deep methods}}	& \multicolumn{4}{c}{\textit{(Deep) GANs based methods}}	\\
			&		&	MRF	&	MWF	&	SSD &	MrFSPS	&	RSLCR	&	BFCN	&	MRNF	&	DGFL  &	BP-GAN	&	cGAN	&	CA-GAN & SCA-GAN	\\
\midrule
	& CUFS     &	68.2 	&	87.0 	&	97.2 &	105.5 	&	106.9 	&	99.7	&	84.5 	&	94.4 &	86.1&	43.2 &	\underline{36.1} &	\textbf{34.2}\\
FID$\downarrow$	& CUFSF    & 70.7	&	86.9	&	75.9	&	87.2 &	126.4	&	123.9	&	--	&	--	&	42.9	&	29.2	&	\underline{19.6}	&	\textbf{18.2}\\
& avg. &  69.5 & 87.0 & 86.6 &  96.8  & 116.6 & 111.8 & -- & -- & 64.5 & 36.2 & \underline{27.8} & \textbf{26.2}  \\

\midrule
	&	CUFS	&	70.4	&	71.4	&	69.6	&	\textbf{73.4} 		&	69.6 	&	69.3 	&	71.4 	&	70.6 &	69.1 	&	71.1 	&	71.3 	&	\underline{71.6} 	\\
FSIM$\uparrow$&	CUFSF	&	69.6	&	70.3	&	68.2	& 68.9	&	66.5 	&	66.2 	&--	&--	&	68.2 &	\underline{72.8} 	&	 72.7	&	\textbf{72.9} 	\\
& avg. &  70.0 & 70.9 & 68.9 & 71.2 & 68.1 & 67.8 & -- & -- & 68.7 & \underline{72.0} & \underline{72.0} & \textbf{72.3}  \\

\midrule
	& CUFS &	88.4 	&	92.3 	&	91.1 &	97.7 &	\underline{98.0} 	&	92.1	&	96.9 	&	\textbf{98.7} &	93.1 	&	95.5 	&	95.8 	&	95.7 \\
NLDA$\uparrow$&	CUFSF	&	45.6	&	73.8	&	70.6&	75.4 	&	75.9 	&	69.8 	&--	&--	&	67.5 	&	\textbf{80.9} 	&	\underline{78.1} 	&	78.0 	\\
& avg. & 67.0 & 83.2 & 80.9 & 86.6 & 87.0 & 81.0 & -- & -- & 85.3 & \textbf{88.2} & \underline{86.9} & \underline{86.8} \\
\bottomrule	
\end{tabular}
\vspace{-0.4cm}
\end{table*}

We first evaluate the effectiveness of our design choices, including (i) using zero-padding instead of resizing in pre-processing, (ii) using face composition (pixel-wise label masks $\mathcal{M}$) as auxiliary input, (iii) the compositional loss $\mathcal{L}_{cmp}$, (iv) the perceptual loss $\mathcal{L}_{vggface}$, and (v) stacked refinement (\textit{stack}). To this end, we construct several model variants and separately conduct photo synthesis and sketch synthesis experiments on the CUFS dataset. Results are shown in Table \ref{tab:ablation} and discussed below.

\subsubsection{Preprocessing} \label{ssec:ablation_preproc} We have used zero-padding and resizing in cGAN, respectively. As shown in Table \ref{tab:ablation}, using zero-padding instead of resizing dramatically improve the performance for both sketch synthesis and photo synthesis, in terms of all the three indices. Such considerable performance improvement demonstrates the significance of keeping image ratios. As a result, we use zero-padding for all the models across all the experiments. 

\subsubsection{Face composition masks} Table \ref{tab:ablation} shows that using the compositional masks as auxiliary input significantly improve the realism of the synthesized face images. Specially, it decreases FID by 2.7 (43.2 $\to$ 40.5) for sketch synthesis and by 36.5 (117.6  $\to$  81.1) for photo synthesis. Besides, this improves the face recognition accuracy from 89.0 to 98.8 for photo synthesis, suggesting that compositional information is essential for photo-based face recognition.

\subsubsection{Compositional loss} Table \ref{tab:ablation} shows that using compositional loss decreases FID by 0.8 (40.5 $\to$ 39.7) for sketch synthesis, while both the image and the compositional masks are used as input. We conduct additional experiments without using the compositional masks as input. Corresponding results show that compositional loss decreases FID by 3.0 (43.2 $\to$ 39.7) for sketch synthesis and by 34.5 (117.6 $\to$ 81.1) for photo synthesis. This highlights our motivation that focusing training on hard components is key. 

\subsubsection{Perceptual loss}
Using the perceptual loss $\mathcal{L}_{vggface}$ significantly reduces the FID values for both sketch synthesis and photo synthesis. Specially, it decreases the FID by 3.6 (39.7 $\to$ 36.1) for sketch synthesis and 39.8 (81.1 $\to$ 41.3) for photo sketch. 
Such comparison results demonstrate that the perceptual loss significantly improve the realism of synthesized face sketches and photos. Besides, the perceptual loss significantly improve the face recognition accuracy by about 20 (79 $\to$ 99) for photo synthesis, on the CUFSF dataset. 

\subsubsection{Stacked refinement}
\label{ssec:exp_stack}
As shown in Table \ref{tab:ablation}, stacked cGANs dramatically decrease the FID of cGAN from 43.2 to 36.6 for sketch synthesis, and from 117.6 to 104.3 for photo synthesis. This suggests that stacked refinement is effective for improving the realism of synthesized images. Likewise, compared to CA-GAN, SCA-GAN further decreases FID by 3.5 (39.7 $\to$ 36.2) for sketch synthesis and by 1.0 (41.3 $\to$ 40.3) for photo synthesis. Besides, SCA-GAN achieves better results than stacked cGANs. 

\begin{figure}
\centering
\includegraphics[width=\linewidth]{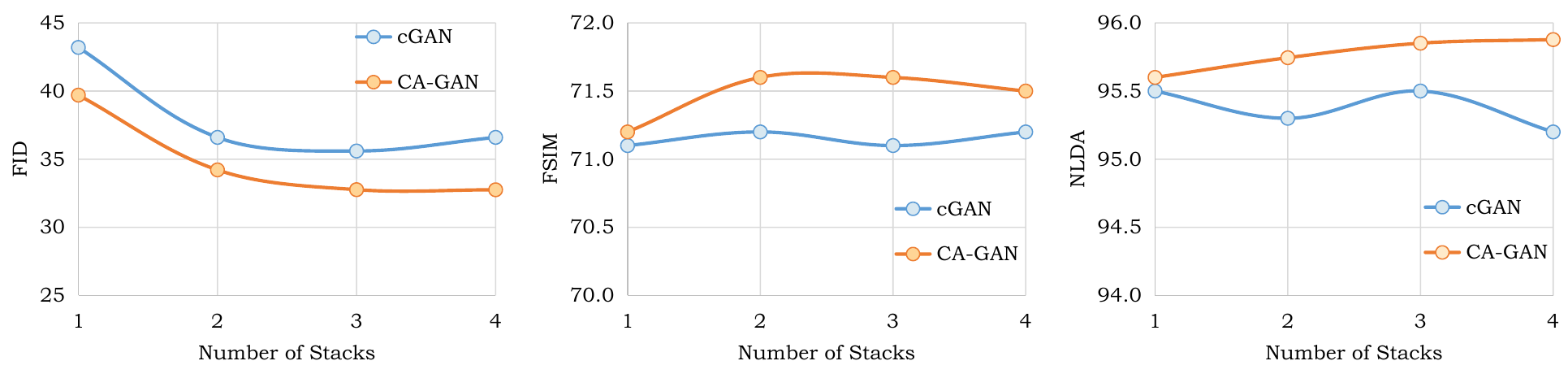}
\vspace{-0.4cm}
\caption{Results of face sketch synthesis while stacking different number of cGANs or CA-GANs, respectively, on the CUFS database.}
\label{fig:exp_stack}
\vspace{-0.4cm}
\end{figure}

In addition, we have evaluated the performance of stacking different number of cGANs or our CA-GANs, respectively. Experiments are conducted on face sketch synthesis on the CUFS database. The corresponding results are shown in Table \ref{tab:exp_stack} and visualized in Fig. \ref{fig:exp_stack}. Here, \textit{one stage} denotes no stacking, and \textit{stack-$k$} denotes stacking $k$ cGANs or CA-GANs, with $k=2, 3, 4$. 
Obviously, the variants of CA-GAN consistently outperform those of cGAN. Specially, by stacking more CA-GANs, we generally generate better sketches, with decreasing FID, slightly increasing NLDA, and nearly invariable FSIM. In contrast, as we stack more cGANs, the FID decreases initially and begins to increase when four cGANs are used. In addition, stacking more than two GANs effects the performance slightly, which is possibly due to the limited number of training examples. However, using a stack of GANs monotonically increase both the computational complexity and model size. We therefore use a stack of two CA-GANs in the rest of this work, unless otherwise specified.

\begin{table}
\centering
\caption{Performance of face sketch synthesis while stacking different number of cGANs or CA-GANs, respectively.}
\label{tab:exp_stack}
\begin{tabular}{c|l|cccc}
\toprule											
	&		&	one stage	&	stack-2	&	stack-3	&	stack-4	\\
\midrule											
\multirow{2}{*}{FID$\downarrow$}	&	cGAN	&	43.2	&	36.6	&	35.6	&	36.6	\\
	&	CA-GAN	&	39.7	&	34.2	&	32.8	&	32.8	\\
\midrule											
\multirow{2}{*}{FSIM$\uparrow$}	&	cGAN	&	71.1	&	71.2	&	71.1	&	71.2	\\
	&	CA-GAN	&	71.2	&	71.6	&	71.6	&	71.5	\\
\midrule											
\multirow{2}{*}{NLDA$\uparrow$}	&	cGAN	&	95.5	&	95.3	&	95.5	&	95.2	\\
	&	CA-GAN	&	95.6	&	95.7	&	95.9	&	95.9	\\
\bottomrule	
\end{tabular}
\vspace{-0.5cm}
\end{table}

To conclude, the proposed approaches significantly improve the realism of synthesized face photos/sketches, with comparative or improved FSIM and NLDA scores. Besides, the performance gains shown in Table \ref{tab:ablation} indicate that the effect of the proposed approaches is at least partly additive.



\subsection{Face Sketch Synthesis}
\label{sec:photo2sketch}

In this section, we compare the proposed methods with a number of existing advanced methods, including MRF \cite{Wang2009Face}, MWF \cite{MWF}, SSD \cite{Song2014Real}, MrFSPS \cite{Peng2016Multiple}, RSLCR\cite{Wang2017RSLCR}, MRNF \cite{Zhang2018Markov}, BFCN \cite{BFCN}, DGFL \cite{Zhu2017Deep}, BP-GAN \cite{wang2017bpgan}, and cGAN \cite{Isola2017Pix2Pix}. Synthesized images of existing methods are released by corresponding authors at: \url{http://www.ihitworld.com/}. 
All these methods and ours follow the same experimental settings. We don't compare with some recently published works, e.g. \cite{Zhang2018IJCAI,Peng2019DeepPGM,Zhang2019TIP,Zhu2019ColGAN,Zhang2019MAL,Zhang2019TCYB}, because experimental settings in these works are specially designed according to their motivations and different from ours.

Table \ref{tab:pfm_sketch} show the results, where "avg." denotes the average value of each criterion across the CUFS dataset and CUFSF dataset. Obviously, our final model, SCA-GAN, significantly decreases the previous state-of-the-art FID by a large margin across both datasets. Besides, CA-GAN obtains the second best FID values. This demonstrates that our methods dramatically improve the realism of the synthesized sketches, compared to existing methods. 

In addition, our methods are highly comparable with previous methods, in terms of FSIM and NLDA. Note that FSIM is designed for evaluating the quality degradation of photos caused by blurring, noise, or compression, and is not suitable for evaluating visual quality of sketches \cite{Wang2016Evaluation}. Besides, Y. Blau and T. Michaeli prove mathematically that “distortion and perceptual quality are at odds with each other” \cite{Blau_2018_CVPR}. In other words, the FSIM value increases, the perceptual quality must be worse. They also show that GANs approaches the perception-distortion bound. Thus the inferiority in the FISM value doesn't mean worse perceptual quality. We report FSIMs here just because it has been widely used in existing works. 

Moreover, the face sketch recognition accuracy is strongly correlated with FSIM, or the fidelity of a synthesized sketch \cite{wang2016training-free}. A lower FSIM score generally corresponds to lower face sketch recognition accuracy. Thus, CA-GAN and SCA-GAN also show slight inferiority in NLDA.
Note that our VGGFace loss network is pre-trained on photos. The perceptual loss is expected to add high-frequency constraints on synthesized sketches and has led to better FID scores (as shown in Table \ref{tab:pfm_sketch}). Besides, as will be seen in Table \ref{tab:pfm_photo}, the perceptual loss significantly improves the performance of face photo synthesis with respect to all the three indices. It is promising to boost the face sketch recognition performance by using a specifically designed loss network. 

\begin{figure}
\centering
\includegraphics[width=\linewidth]{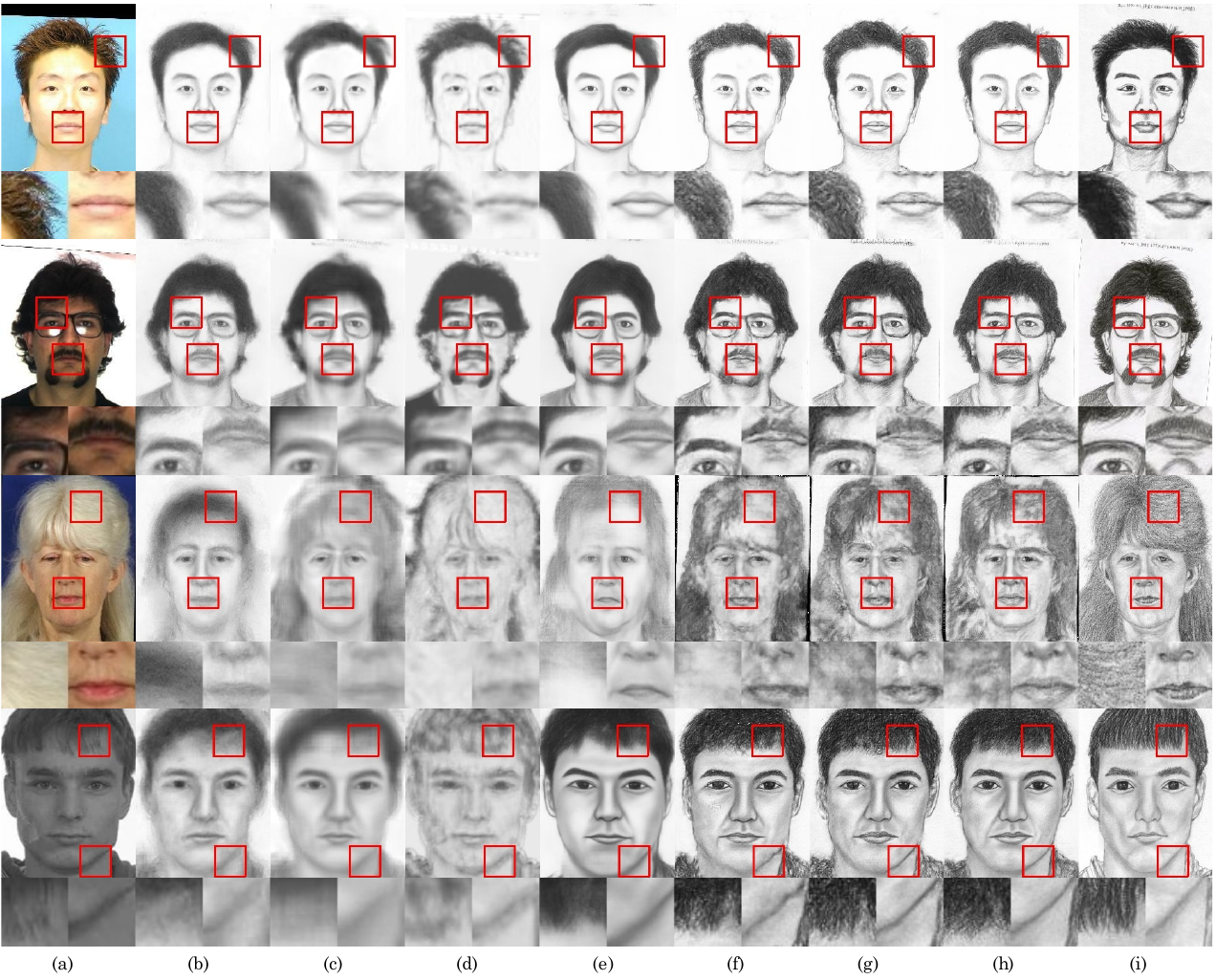}
   \vspace{-0.8cm}
\caption{Examples of synthesized face sketches on the the CUFS dataset and the CUFSF dataset.
(a) Photo, (b) MrFSPS \cite{Peng2016Multiple}, (c) RSLCR\cite{Wang2017RSLCR}, (d) BFCN \cite{BFCN}, (e) BP-GAN \cite{wang2017bpgan}, (f) cGAN \cite{Isola2017Pix2Pix}, (g) CA-GAN, (h) SCA-GAN, and (i) Sketch drawn by artist. From top to bottom, the examples are selected from the CUHK student dataset \cite{Ref2}, the AR dataset \cite{Ref23}, the XM2VTS dataset \cite{Ref24}, and the CUFSF dataset \cite{Ref32}, sequentially.}
\label{fig:sketch_cmp_tra}
   \vspace{-0.4cm}
\end{figure}

Fig. \ref{fig:sketch_cmp_tra} presents some synthesized face sketches from different methods on the CUFS dataset and the CUFSF dataset. Due to space limitation, we only compare to several advanced methods here. Obviously, MrFSPS, RSLCR, and BFCN yield serious blurred effects and great deformation in various facial parts. In contrast, GANs based methods can generate sketch-like textures (e.g. hair region) and shadows. However, BP-GAN yields over-smooth sketch portraits, and cGAN yields deformations in synthesized sketches, especially in the mouth region. Notably, CA-GAN alleviates such artifacts, and SCA-GAN almost eliminates them. 

To conclude, both the qualitative and quantitative evaluations shown in Table \ref{tab:pfm_sketch} and Fig. \ref{fig:sketch_cmp_tra} demonstrate that both CA-GAN and SCA-GAN are capable of generating quality sketches. Specially, our methods achieve significantly gain in realism of synthesized sketches over previous state-of-the-art methods. Besides, our methods perform on par with previous state-of-art methods in term of FSIM and face sketch recognition accuracy.

\subsection{Face Photo Synthesis}
\label{sec:sketch2photo}

We exchange the roles of the sketch and photo in the proposed model, and evaluate the face photo synthesis performance. 
%
%
To our best knowledge, only few methods have been proposed for face photo synthesis. Here we compare the proposed method with one advanced method: MrFSPS \cite{Peng2016Multiple}. As shown in Table \ref{tab:pfm_photo}, 
both CA-GAN and SCA-GAN significantly outperform existing methods in general, according to all these criteria. In other words, SCA-GAN and CA-GAN approach the perception-distortion bound. Specially, CA-GAN reduces previous best (average) FID from 60.9 to 32.7 for photo synthesis; and SCA-GAN achieves better performance than CA-GAN. In addition, both CA-GAN and SCA-GAN considerably improve the face recognition accuracy by about 5 and 20 percent on the CUFS dataset and CUFSF dataset, respectively. 
Such considerable performance improvement demonstrates that our method can produce both perceptually realistic and identity-preserving face photos. 

\begin{figure}
\centering
\includegraphics[width=0.67\columnwidth]{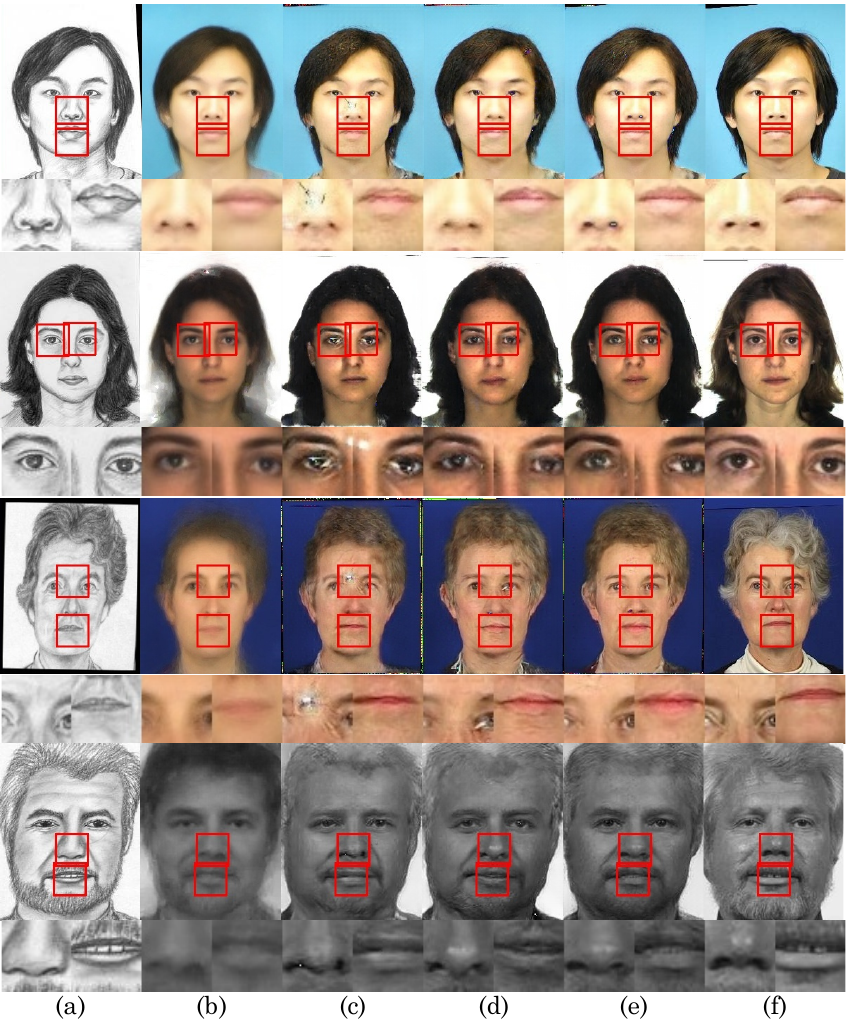}
   \vspace{-0.4cm}
\caption{Examples of synthesized face photos. (a) Sketch drawn by artist, (b) MrFSPS \cite{Peng2016Multiple}, (c) cGAN, (d) CA-GAN, (e) SCA-GAN, and (f) ground-truth photo. From top to bottom, the examples are selected from the CUHK student dataset \cite{Ref2}, the AR dataset \cite{Ref23}, the XM2VTS dataset \cite{Ref24}, and the CUFSF dataset \cite{Ref32}, sequentially.}
\label{fig:sketch2photo}
\end{figure}

Fig.\ref{fig:sketch2photo} shows examples of synthesized photos. Obviously, results of MrFSPS are heavily blurred. Besides, there is serious degradation in the synthesized photos by using cGAN. In contrast, the photos generated by CA-GAN or SCA-GAN consistently show considerable improvement in perceptual quality. Results of CA-GAN and SCA-GAN express more natural colors and details. Recall the quantitative evaluations shown in Table \ref{tab:pfm_photo}, we can safely draw the conclusion that our methods are capable of generating natural face photos while preserving the identity of the input sketch. 

\begin{table}
\centering
\caption{Performance on face photo synthesis on the CUFS and CUFSF datasets. The best and second best results in each row are shown in \textbf{boldface} and \underline{underline} format, respectively.  $\downarrow$ indicates lower is better, while $\uparrow$ higher is better.}
\footnotesize
\label{tab:pfm_photo}
\begin{tabular}{c|c|cccc}
\toprule
Criterion	&	Dataset	&	MrFSPS & cGAN	&	CA-GAN	&	SCA-GAN	\\
\midrule									
	&	CUFS	&	92.0 &	88.7	&	\underline{41.3}	&	\textbf{40.3} 	\\
FID$\downarrow$	&	CUFSF	&	95.6 &	33.1	&	\underline{24.4}	&	\textbf{20.6}	\\
	& avg. & 93.8 & 60.9 & \underline{32.7} & \textbf{30.5} \\
\midrule
	&	CUFS	&	\textbf{80.3} &	76.2 	&	78.6 	&	\underline{79.5} 	\\
FSIM$\uparrow$&	CUFSF	& 79.3	&	79.5 	&	 \underline{83.7} 	&	\textbf{84.5} 	\\
	& avg.		&	79.8 & 77.8 & \underline{81.1}  & \textbf{82.0} \\
\midrule
	&	CUFS		&	96.7 &	94.8 	&	\underline{98.5} 	&	\textbf{99.4} 	\\
NLDA$\uparrow$&	CUFSF	&	59.4 &	77.5 	&	\underline{99.8}  	&	\textbf{ 99.9 } 	\\
	& avg. &  78.2 & 86.1 &  \underline{99.2} & \textbf{99.7} \\
\bottomrule							
\end{tabular}
\vspace{-0.5cm}
\end{table}

\subsection{Robustness Evaluation}
\label{sec:crossdataset}

To verify the generalization ability of the learned model, we apply the model trained on the CUFS training dataset to faces in the wild.

\subsubsection{Lighting and Pose Variations} 
\label{sec:robust_light_pose}
\begin{figure}
\centering
\includegraphics[width=0.5\columnwidth]{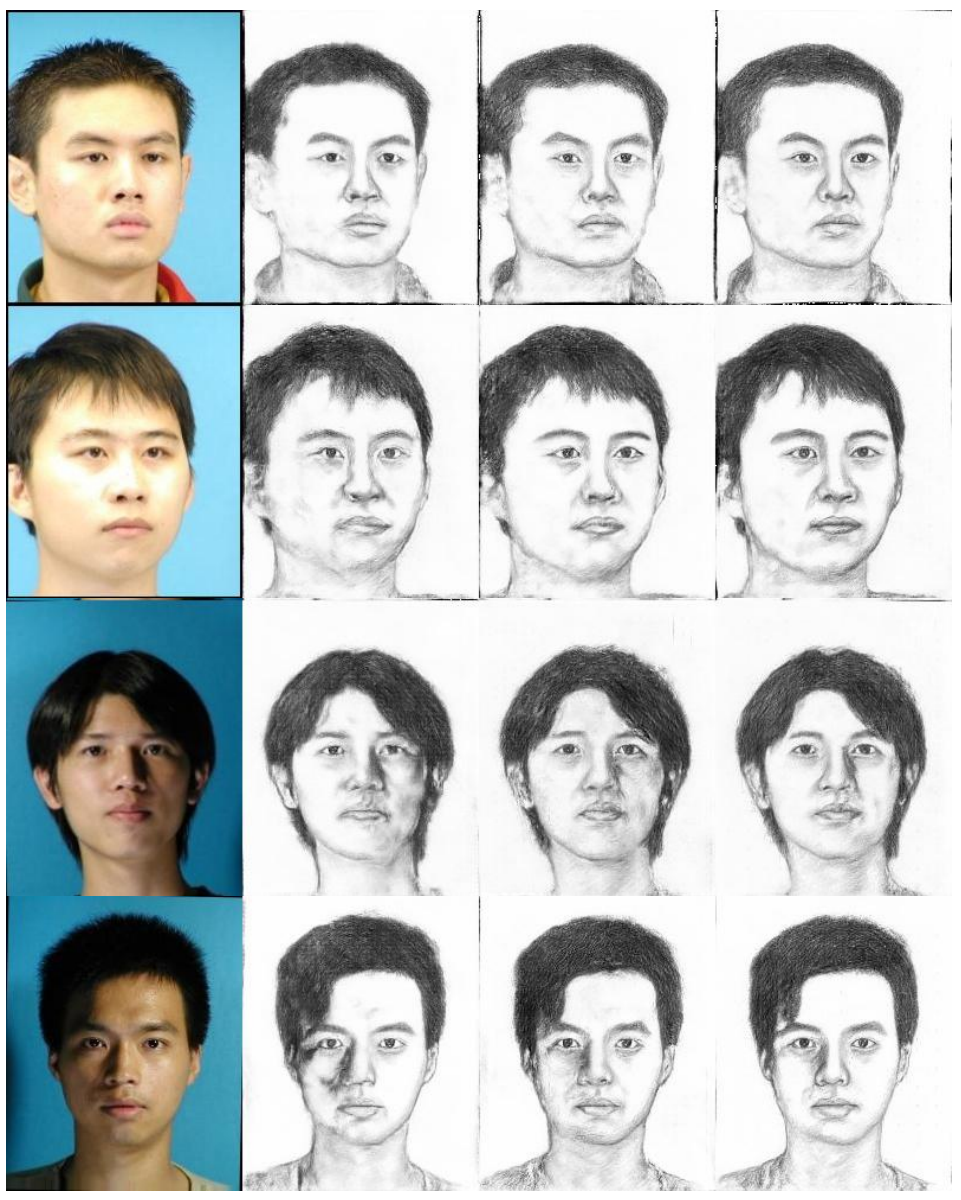}
   \vspace{-0.4cm}
\caption{Robustness to lighting and pose variations. (a) Photo, (b) cGAN, (c) CA-GAN, and (d) SCA-GAN. }
\label{fig:crossdata_sketch}
   \vspace{-0.3cm}
\end{figure}

We first apply the learned models to a set of face photos with lighting variation and pose variation. Fig. \ref{fig:crossdata_sketch} shows some synthesized results from cGAN, CA-GAN, and SCA-GAN. Clearly, results of cGAN exists blurring and inky artifacts over dark regions. In contrast, the photos produced by CA-GAN and SCA-GAN show less artifacts and express improved details over the eye and mouth regions. Besides, results of SCA-GAN show the best quality. 

\subsubsection{Face photo-sketch synthesis of national celebrities} 
\label{sec:celeb}

We further test the learned models on the photos and sketches of national celebrities. 
All these photos and sketches are downloaded from the web and adopted as input. These images contain different lighting conditions and backgrounds compared with the images in the training set. 

\begin{figure}
\centering
\includegraphics[width=0.48\columnwidth]{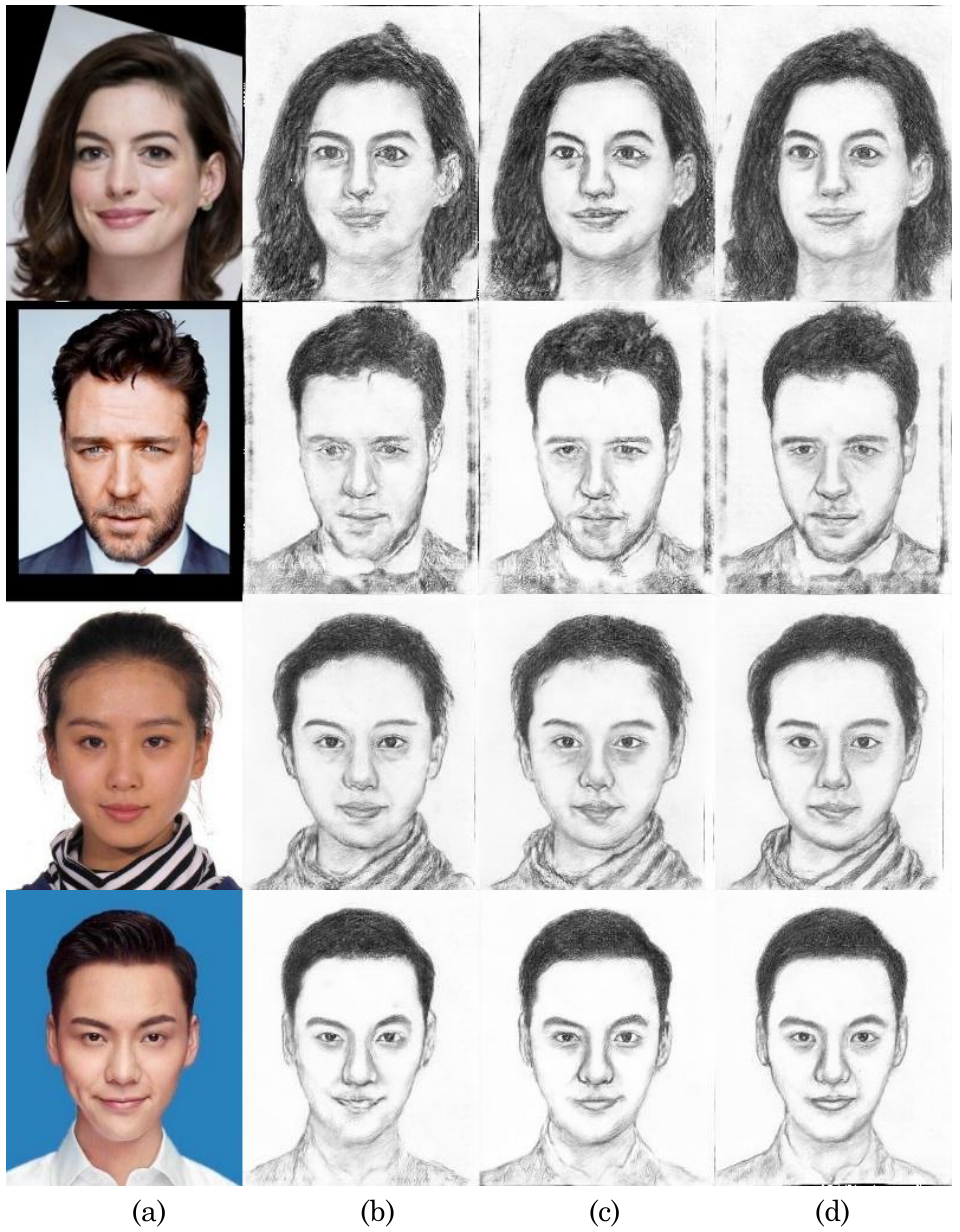}
\includegraphics[width=0.48\columnwidth]{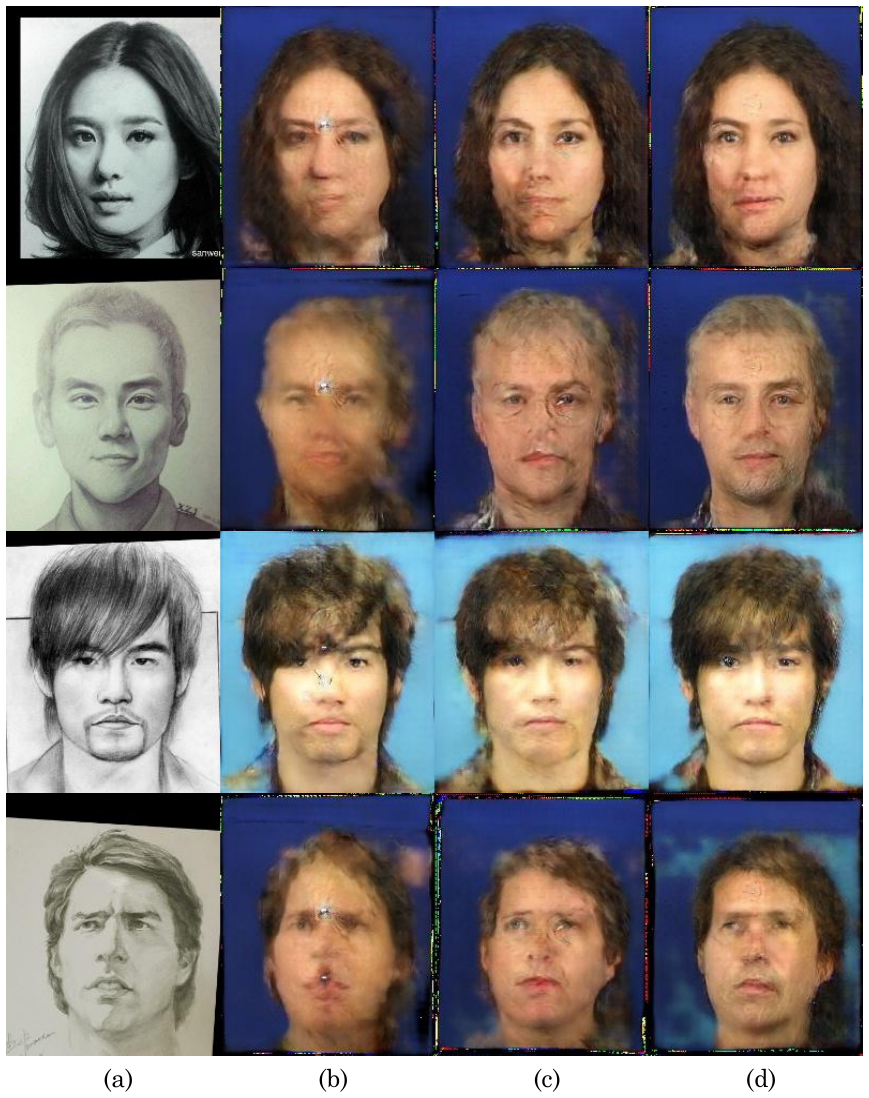}
  \vspace{-0.4cm}
\caption{Face photo-sketch synthesis results of national celebrities. (a) Input sketch/photo, (b) cGAN, (c) CA-GAN, (d) SCA-GAN. }
\label{fig:celeb_ps}
   \vspace{-0.4cm}
\end{figure}

Fig. \ref{fig:celeb_ps} shows the synthesized sketches and photos. Obviously, our results express more natural textures and details than cGAN, for both sketch synthesis and photo synthesis. Both CA-GAN and SCA-GAN show outstanding generalization ability in the sketch synthesis task. The synthesized photos here are dissatisfactory. This might be due to the great divergence between the input sketches in terms of textures and styles. It is necessary to further improve the generalization ability of the photo synthesis models.

\subsection{Analysis of the Training Procedure}
\label{sec:loss}

\begin{figure}
\centering
\includegraphics[width=1\linewidth]{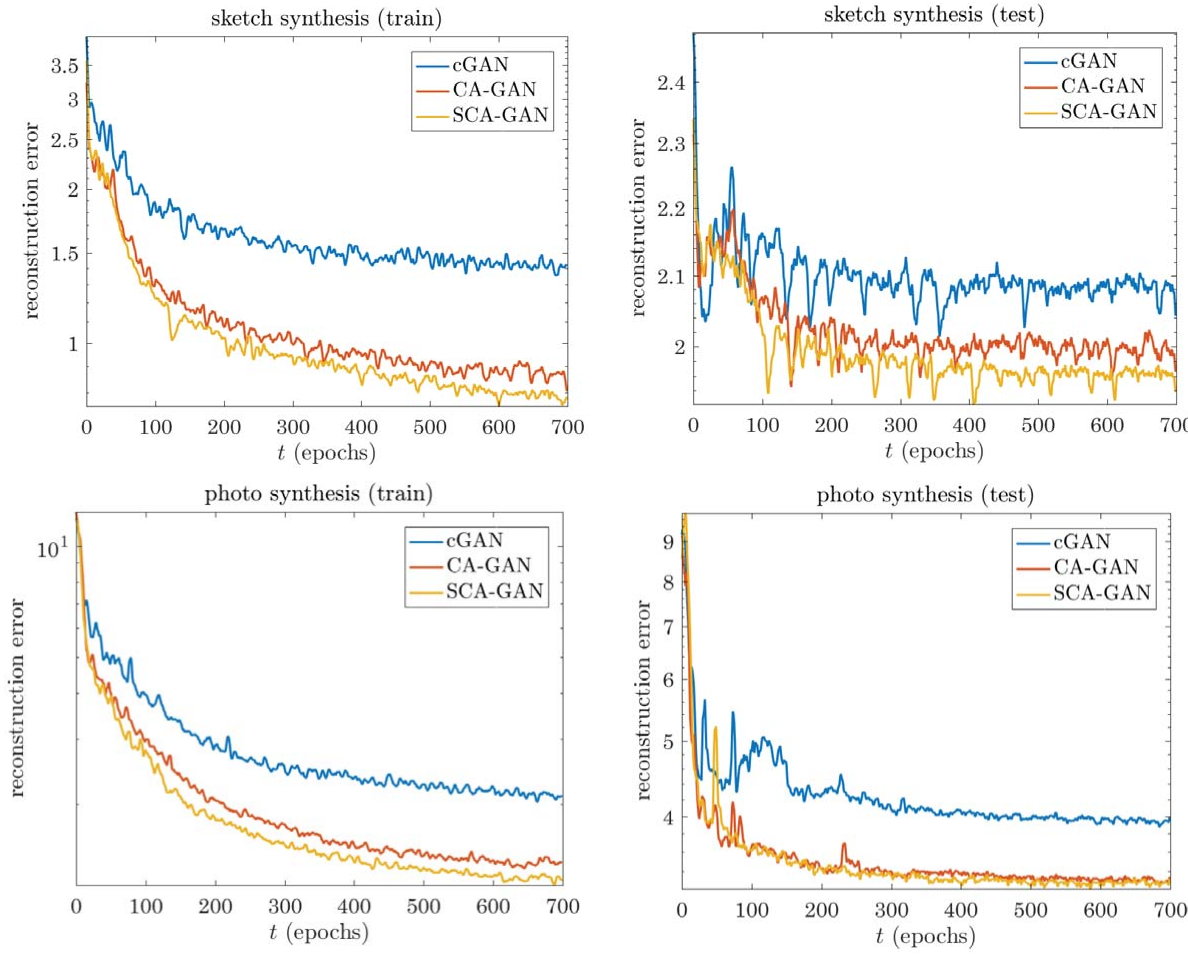}
\vspace{-0.8cm}
\caption{Reconstruction errors about sketch synthesis and photo synthesis during the training process, on the CUFS dataset. \textit{train} denotes the reconstruction error on the training set, and \textit{test} that on the testing set.}
\label{fig:loss}
\vspace{-0.6cm}
\end{figure}

We finally analyse the training procedure of cGAN \cite{Isola2017Pix2Pix}, CA-GAN, and SCA-GAN. To this end, we conduct sketch synthesis and photo synthesis experiments on the CUFS dataset, respectively.
During training, after every epoch, we apply the learned model to the training/testing subset and calculate the corresponding reconstruction errors (Global L1 loss) on each subset.
Fig. \ref{fig:loss} shows the reconstruction errors, where \textit{train} denotes the reconstruction error on the training set, and \textit{test} that on the testing set. For clarity, these curves are plot on semilog coordinate. 

Obviously, there are relatively larger fluctuations in loss curves of cGAN, in contrast to those of CA-GAN and SCA-GAN. Besides, the reconstruction errors of both CA-GAN and SCA-GAN drop faster initially and are lower than that of cGAN. Finally, SCA-GAN shows slight superiority over CA-GAN in terms of the reconstruction errors. These observations demonstrate that the proposed approaches considerably speed the training up and stabilize it. 

\subsection{Summary}
\label{ssec:summary}

In this part, we briefly summary the conclusions we could draw from the experimental results.  
\begin{itemize}
\item First, our model is capable of generating both visually realistic and identity-preserving face sketches/photos over a wide range of challenging data. Specially, our model significantly decrease previous state-of-the-art FID from 36.2 to 26.2 for sketch synthesis, and from 60.9 to 30.5 for photo synthesis; 

\item Second, both using face composition as supplemental input and training with the compositional loss increase the realism of synthesized sketches/photos;

\item Third, the perceptual loss significantly improves the quality of synthesized photos, but contributes little to face sketch synthesis. It is promising to using a face sketch recognition network to boost the quality of synthesized sketches;

\item Forth, stacking CA-GANs generally improve the quality of synthesized face photos/sketches;

\item Fifth, the proposed approaches speed the training up and stabilize it; and

\item Finally, both CA-GAN and SCA-GAN are of significantly improved generalization ability, especially for face sketch synthesis. It is challenging to produce perceptually comfortable photos from face sketches in the wild. 

\end{itemize}

\section{Conclusion}
\label{sec:conclusion}

In this paper, we propose a novel composition-aided generative adversarial network (CA-GAN) for face photo-sketch synthesis. Our approach dramatically improves the realism of the synthesized face photos and sketches over previous state-of-the-art methods. We hope that the presented approach can support applications of other image generation problems. Besides, it is essential to develop models that can handle photos/sketches with great variations in head poses, lighting conditions, and styles. To this end, exploring hierarchical deep features \cite{Yu2019PAMI, Yu2019TNNLS} and using multi-task learning \cite{Yu2018TIE} might be promising solutions. Exciting work remains to be done to qualitatively evaluate the quality of synthesized sketches and photos. Here the reliability of FID is indeterminate, because the dimension of deep features is dramatically higher than the number of photo/sketh samples. Using dimension reduction \cite{LIU2019DLRPP} before computating FID might be a solution. Finally, it is meaningful to apply face photo-sketch synthesis algorithms to practices, such as image privacy protection \cite{Yu2017TIFS}. 

%
%
%
%




\bibliographystyle{IEEEtran}
\bibliography{IEEEabrv,ref}
\end{document}